# A generalised pre-training strategy for deep learning networks in semantic segmentation of remotely sensed images

Yuan Fang[1], Yuanzhi Cai[2], Jagannath Aryal[3], Qinfeng Zhu[1], Hong Huang[1], Cheng Zhang[1], and Lei Fan[1]*

[1] Department of Civil Engineering, Xi'an Jiaotong-Liverpool University, Suzhou 215000, China
[2] CSIRO Mineral Resources, Kensington, WA 6151, Australia
[3] Department of Infrastructure Engineering, The University of Melbourne, Melbourne VIC 3010, Australia

* Corresponding Author: lei.fan@xjtlu.edu.cn

**Abstract.** In the segmentation of remotely sensed images, deep learning models are typically pre-trained using large image databases like ImageNet before fine-tuned on domain-specific datasets. However, the performance of these fine-tuned models is often hindered by the large domain gaps (i.e., differences in scenes and modalities) between ImageNet's images and remotely sensed images being processed. Therefore, many researchers have undertaken efforts to establish large-scale domain-specific image datasets for pre-training, aiming to enhance model performance. However, establishing such datasets is often challenging, requiring significant effort, and these datasets often exhibit limited generalizability to other application scenarios. To address these issues, this study introduces a novel yet simple pre-training strategy designed to guide a model away from learning domain-specific features in a pre-training dataset during pre-training, thereby improving the generalisation ability of the pre-trained model. To evaluate the strategy's effectiveness, deep learning models are pre-trained on ImageNet and subsequently fine-tuned on four semantic segmentation datasets with diverse scenes and modalities, including iSAID, MFNet, PST900 and Potsdam. Experimental results show that the proposed pre-training strategy led to state-of-the-art accuracies on all four datasets, namely 67.4% mIoU for iSAID, 56.9% mIoU for MFNet, 84.22% mIoU for PST900, 91.88% mF1 for Potsdam. This research lays the groundwork for developing a unified foundation model applicable to both computer vision and remote sensing applications.

## 1 Introduction

Semantic segmentation is essential in remote sensing (RS) applications, impacting areas from land use [1] and land cover classification [2, 3] to forest monitoring [4] and urban scene understanding [5, 6]. Leveraging deep learning models, especially those pre-trained on large image datasets like ImageNet [7], has become a standard approach to enhance training efficiency and accuracy in these applications [8, 9]. These models are typically pre-trained on ImageNet and then fine-tuned on domain-specific datasets to adapt to the nuances of remote sensing imagery. Remote sensing images, however, often include scenes and feature channels, such as different spectral bands and modalities, not covered by ImageNet [10]. This discrepancy introduces significant domain gaps that affect the accuracy when fine-tuning, driving the need for domain-specific training datasets [11]. Despite the advantages of using dedicated RS datasets, their development is hampered by challenges in acquiring large-scale, labeled datasets specific to various RS applications. This limitation has led researchers to explore self-supervised learning techniques, such as contrastive learning (CL) and masked image modeling (MIM), which utilize unlabeled RS images to build models without extensive manual annotation efforts [12]. While this approach addresses some challenges, it also introduces issues related to the generalization capabilities of the models trained exclusively on domain-specific data.

The main contributions of this study are summarised as follows:
1) We propose a novel pre-training strategy named CSP, which greatly enhances the generalisation ability of pre-trained models for semantic segmentation of images.

2) CSP enables the creation of pre-trained models using only the ImageNet dataset while achieving SOTA fine-tuning accuracies on four distinct semantic segmentation datasets characterised by diverse scenes and image modalities, under CNN or transformer architectures.

## 2 Related Work

The exploration of domain adaptation techniques reveals that spectral differences between training and application datasets significantly constrain model performance across diverse scenarios [13]. Prior work in this area has highlighted that traditional domain-specific pre-training approaches, though beneficial in narrow applications, fail to generalize across different contexts. This limitation is evident in RS applications where models trained on dedicated datasets excel in their target domain but underperform when applied to new domains or modal imagery [12]. Additionally, the reliance on large annotated datasets poses a non-trivial constraint in terms of resource demands and scalability.

To address these issues, researchers have advocated for the creation of more versatile and adaptable pre-trained models. The introduction of models that do not focus solely on domain-specific features during pre-training represents a promising avenue [13]. Techniques that neutralize the bias towards learning domain-specific spectral features (e.g., colors specific to a dataset) during pre-training can enhance a model's ability to generalize. This approach encourages models to prioritize learning spatial and other relevant features critical for semantic segmentation, which are more universally applicable across various domains [13].

In continuance, we propose a novel pre-training strategy named Channel Shuffling Pre-training (CSP) aimed at breaking the norm of domain-dependent learning by fostering a model's ability to generalize across both traditional computer vision and remotely sensed imagery scenarios. CSP involves manipulating the input channels during the pre-training phase to de-emphasize the learning of domain-specific spectral characteristics, thereby focusing on more generic image features beneficial for semantic segmentation tasks in multiple contexts. This strategy's effectiveness is validated through extensive fine-tuning experiments across diverse datasets including RGB, multispectral, and multimodal images from different RS applications [14–16]. To the best of our knowledge, it is the first time that pre-trained models can simultaneously adapt to RGB, multispectral, and multimodal images from both RS and computer vision scenes. This potentially eliminates the necessity of collecting large-scale domain-specific datasets and lays the groundwork for developing a unified foundation model.

## 3 Methodology

### 3.1 Dataset

The ImageNet 2012 classification dataset (ImageNet-1K) is used for pre-training, which consists of 1000 object classes, 1.28 million training images, and 50 k validation images. To test the effectiveness of the proposed CSP strategy, four semantic segmentation datasets (i.e., iSAID, MFNet, PST900 and Potsdam) collected using different sensors in diverse scenarios are used. The key characteristics of these datasets are summarised in Table 1.

**Table 1.** Summary of four semantic segmentation datasets used

| Dataset | Number of images | | | Classes | Image resolution | Modalities | Scenarios |
|---------|----------|------------|------|---------|------------|------------|-----------|
|         | Training | Validation | Test |         |            |            |           |
| iSAID   | 1411     | 458        | 937  | 16      | 800×800 to 4000×13000 | R, G, B | RS aerial images |
| MFNet   | 784      | 392        | 393  | 9       | 480×640    | R, G, B, T | Autonomous driving |
| PST900  | 597      | -          | 288  | 5       | 720×1280   | R, G, B, T | Subterranean images |
| Potsdam | 24       | -          | 14   | 6       | 6000×6000  | R, G, B, IR, | RS aerial images |

### 3.2 CSP Strategy

In our proposed CSP strategy, the dataset (e.g., ImageNet) used for pre-training a model can have a different number of image channels from the dataset used for subsequent model fine-tuning. Specifically, to obtain a pre-trained model with $n$ input channels, the CSP strategy (see Fig. 1) first duplicates the $m$ channels of original pre-training images $x$ times and concatenates them, forming temporary input images with $x * m$ channels. The integer $x$ is chosen such that $x * m$ is equal to or slightly greater than $n$. Subsequently, channel random shuffling is performed on temporary input images ($x * m$ channels), and the first $n$ channels are selected as final input images for pre-training the model. It is notable that channel shuffling removes the fixed channel order from the original input image dataset, thus potentially preventing the pre-trained model from learning specific spectral features.

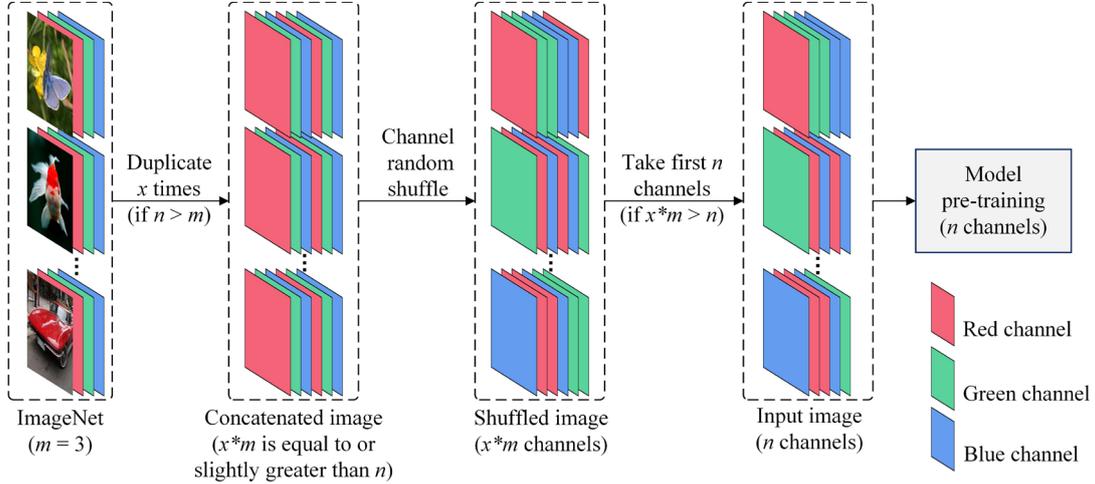

**Fig. 1.** Overall architecture of the channel shuffling pre-training (CSP) strategy.

In this study, to demonstrate the effectiveness of the CSP strategy, we use only the ImageNet dataset ($m = 3$) to pre-train multiple models with 3, 4, and 5 input channels. The corresponding CSP strategies are denoted as CSP-3, CSP-4, and CSP-5, respectively. Fig. 1 shows an example of the CSP-5 strategy where ImageNet ($m = 3$) is used to pre-train a five-channel model ($n = 5$). In this example, the three channels of the input RGB images are duplicated 2 times ($x = 2$). After channel shuffling of the resulting 6-channel images, the first 5 channels are used for model pre-training.

### 3.3 Training Setting

Given their proven performance in previous research [17–19], ConvNeXt and Swin Transformer are selected as the network architectures for pre-training models in this study. For fair comparison, the same decoder UperNet [20] is used for all fine-tuning experiments. All experiments are conducted using the Pytorch framework and rely on two Nvidia GeForce RTX 3090 GPUs.

Table 2 summarises the pre-training and fine-tuning settings used in this study. These settings largely mirror their original implementation, with the following exceptions. Due to limited Graphics Processing Unit (GPU) memory, the batch size for pre-training is reduced to 128, with a corresponding reduction in the learning rate. For pre-processing Potsdam and iSAID datasets, the same strategy as previous studies [21] is adopted, involving the cropping of original images into 512×512 and 896×896 patches, respectively. For Potsdam, the class of "clutter" were ignored for both training and testing. Random cropping, random resize, and random horizontal flipping are used for data augmentation [22].

## 4 Experiments and Results

### 4.1 Experiment Setup

A series of experiments were conducted on the CSP strategy using ConvNeXt-T and Swin-T (the tiny version of ConvNeXt and Swin, respectively) for comparisons against original pre-training strategy. In this study, the performances of ConvNeXt-T and Swin-T pre-trained with their original implementations were used as baselines. This pre-training setting is denoted as IM-RGB. For the three-channel iSAID dataset, the four-channel MFNet and PST900 datasets and the five-channel Potsdam dataset, we used the CSP-3, CSP-4, and CSP-5 strategies, respectively, in the ConvNext and Swin Transformer models. To measure performance, segmentation maps are obtained using single-scale testing, with accuracy evaluations based on the following metrics, namely mean F1 score (mF1) and mean intersection over union (mIoU).

**Table 2.** Detailed pre-training and fine-tuning setting

| Pre-training | | | | |
|---|---|---|---|---|
| Patch size | 224×224 | | | |
| Total training epochs | 300 | | | |
| Batch size | 128 | | | |
| Hyperparameters | $\beta_1$=0.9; $\beta_2$=0.999; $\epsilon$ = 1e$^{-8}$ | | | |
| Optimizer | AdamW | | | |
| Weight decay | 0.05 | | | |
| Loss function | Cross entropy | | | |
| Learning rate | 0.000125 | | | |
| Learning rate schedule | Linear warmup for 20 epochs with a start factor: 1e$^{-3}$, and then use cosine decay policy | | | |
| Data augmentation | Random cropping, random resize (0.5, 0.75, 1.0, 1.5, 1.75) and random horizontal flipping | | | |
| Fine-tuning | | | | |
| Dataset | Potsdam | iSAID | MFNet | PST900 |
| Patch size | 512×512 | 896×896 | 512×512 | 512×512 |
| Total training iterations | 10 k | 80k | 10 k | 10 k |
| Batch size | 16 | 8 | 16 | 16 |
| Hyperparameters | $\beta_1$=0.9; $\beta_2$=0.999 | | | |
| Optimizer | AdamW | | | |
| Weight decay | 0.05 for ConvNeXt; 0.01 for Swin | | | |
| Loss function | Cross entropy | | | |
| Learning rate | 0.0001 for ConvNeXt; 0.00006 for Swin | | | |
| Learning rate schedule | Linear warmup for 1500 iterations with a start factor of 1e$^{-6}$, and then use polynomial decay policy with a power of 1.0 | | | |
| Data augmentation | Random cropping, random resize (0.5, 0.75, 1.0, 1.5, 1.75) and random horizontal flipping | | | |

## 4.2 Pre-training on ImageNet-1K

Table 3 summarises the classification accuracies of different pre-trained models on the ImageNet-1K validation set. The Top-1 classification accuracies of the models using the CSP strategy were, on average, 3.76% and 1.42% lower than those of the baseline models, respectively. These decreases in classification accuracies were expected since the CSP strategy was intentionally designed to prevent the model from utilizing spectral information. Specifically, the strategy dilutes the model's ability to exploit dataset-specific color patterns, promoting a more substantial focus on generic, spatial, and structural features which are crucial for domain-agnostic applications. Thus, while this approach anticipates reduced performance on tasks heavily dependent on spectral information, it is strategically devised to bolster the model's generalization capabilities across diverse visual recognition scenarios, potentially enhancing its applicability in more generalized settings beyond standard benchmarks.

**Table 3.** Classification accuracies on ImageNet-1K validation set with different pre-trained models and spectral channel sequences

| Networks | Pretrain | RGB Input | | BGR Input | |
|---|---|---|---|---|---|
| | | Top-1 | Top-5 | Top-1 | Top-5 |
| ConvNeXt-T [50] | IM-RGB | **82.10** | **95.89** | - | - |
| | CSP-3 | 80.53 | 95.16 | - | - |
| | CSP-4 | 80.63 | 95.27 | - | - |
| | CSP-5 | 79.95 | 94.62 | - | - |
| Swin-T [51] | IM-RGB | **81.18** | **95.61** | 72.30 | 91.58 |
| | CSP-3 | 79.32 | 94.68 | 79.22 | 94.73 |
| | CSP-4 | 79.81 | 94.72 | - | - |
| | CSP-5 | 80.14 | 94.85 | - | - |

In our study, we also tested Swin-T models on images with reversed channel sequences (BGR), and the results from Table 3 indicate that the CSP strategy resulted in minimal changes in classification accuracy due to spectral perturbations, as shown in Fig. 2's activation maps. In contrast, the Swin-T model trained with the standard IM-RGB strategy exhibited notable sensitivity to these perturbations, leading to misclassifications, as depicted for three images in Fig. 2. This confirms the effectiveness of CSP in reducing dependency on spectral features.

Although the CSP-trained Swin-T had a slightly lower overall accuracy compared to IM-RGB, it performed better in several specific classes, as detailed in Table 4. For classes involving artificially colored objects such as vases and desks, or those challenging to separate from their backgrounds like echidnas and American chameleons, CSP's resistance to spectral changes and emphasis on spatial features proved beneficial. For instance, the third row of Fig. 2 illustrates how CSP-trained Swin-T correctly identified an American chameleon but the IM-RGB model misidentified it as a green lizard. The CSP strategy's focus on global features also likely contributed to improved semantic segmentation accuracy during fine-tuning.

**Table 4.** Examples of advantageous classes on ImageNet-1K of Swin-T trained by CSP-3

| Classes | CSP-3 | | IM-RGB | |
|---|---|---|---|---|
| | RGB | BGR | RGB | BGR |
| Echidna | 64% | 64% | 44% | 42% |
| Consomme | 80% | 80% | 62% | 46% |
| American chameleon | 86% | 84% | 68% | 42% |
| Vase | 64% | 60% | 50% | 50% |
| Alp | 72% | 74% | 56% | 50% |
| Desk | 68% | 68% | 52% | 54% |

### 4.3 Fine-tuning on iSAID, MFNet, PST900 and Potsdam

This section reports the performances of fine-tuned models on iSAID, MFNet, PST900 and Potsdam. These fine-tuned models are based on prior pre-training using both CSP and IM-RGB, facilitating a comparative evaluation of the performance enhancement by the CSP strategy and its generalizability across different datasets.

**A. Quantitative results.** The segmentation results for the iSAID dataset, detailed in Table 5, attest to the effectiveness of the CSP framework. Since this dataset contains only the R-G-B channels, the CSP-3 strategy was employed. Our CSP-3 strategy resulted in mIoU improvements of 0.8% for ConvNext-T and 1.6% for Swin-T compared to the IM-RGB strategy.

Our strategy also showed significant improvements in the MFNet and PST900 datasets, both of which consist of R-G-B-T spectral band combinations, as shown in TABLE 6 and TABLE 7. Experiments conducted using ConvNeXt-T and Swin-T models revealed that, on average, the CSP-4 strategy improved the mIoU scores by 6.6% and 13.83% over the IM-RGB scores in MFNet and PST900, respectively. Fig. 3 illustrates the dramatic changes in illumination conditions of scenes in PST900, where the colors of the objects often closely resembled one another and the background. The exceptional performance under such challenging conditions underscores the potential applicability of the CSP framework in autonomous driving applications, especially under extreme weather conditions.

**Table 5.** Semantic segmentation accuracies on iSAID validation set (%)

| Networks | Pretrain | mIoU | IoU per class | | | | | | | | | | | | | |
|---|---|---|---|---|---|---|---|---|---|---|---|---|---|---|---|---|
| | | | Ship | ST | BD | TC | BC | GTF | Bridge | LV | SV | HC | SP | RA | SBF | Plane | Harbor |
| ConvNeXt-T [17] | IM-RGB | 66.6 | **72.8** | 75.0 | 77.5 | **89.0** | **67.4** | 59.3 | 42.0 | **66.0** | **51.9** | **38.0** | 50.7 | 57.6 | 73.4 | 85.2 | **60.1** |
| | CSP-3 | **67.4** | 71.6 | **75.5** | **79.3** | 88.9 | 66.1 | **59.7** | **42.6** | 65.2 | 51.8 | 36.9 | **51.0** | **71.6** | **76.4** | **85.2** | 58.4 |
| Swin-T [18] | IM-RGB | 64.6 | 69.2 | **76.5** | 74.1 | 69.9 | 56.3 | **60.1** | 41.9 | 62.3 | **51.6** | **44.7** | 45.8 | 64.5 | 75.9 | **85.7** | 56.7 |
| | RSP [23] | 64.1 | 67.0 | 74.6 | 73.7 | 70.7 | 59.0 | 60.1 | **44.3** | 62.0 | 50.6 | 37.6 | 46.8 | 64.9 | **76.2** | 85.2 | 53.8 |
| | CSP-3 | **66.2** | **71.1** | 73.7 | **76.7** | **88.5** | **61.8** | 59.9 | 38.6 | **65.5** | 50.7 | 39.7 | **49.1** | **67.4** | 75.4 | 84.4 | **58.1** |

ST: storage tank. BO: baseball diamond. TC: tennis court. BC: baseball court. GTF: ground track field. LV: large vehicle. SV: small vehicle. HC: helicopter. SP: swimming pool. RA: roundabout. SBF: soccer ball field.

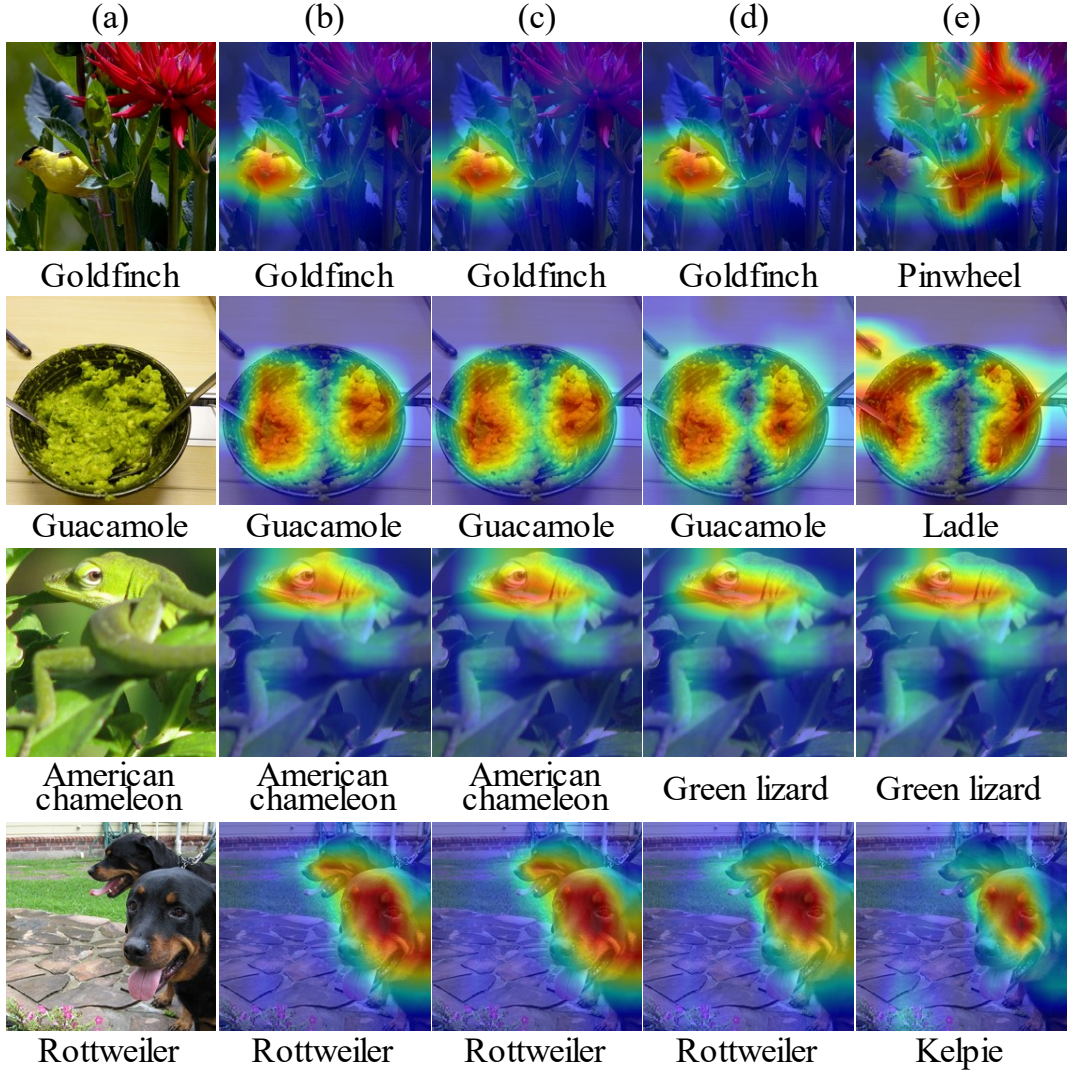

**Fig. 2.** Activation maps and prediction results (underneath) of Swin-T models on different scenes: (a) Original images, (b) CSP-3 with RGB input, (c) CSP-3 with BGR input, (d) IM-RGB with RGB input, (e) IM-RGB with BGR input.

**Table 6.** Semantic segmentation accuracies on MFNet test set (%)

| Networks | Pretrain | mIoU | IoU per class | | | | | | | |
| | | | Car | Person | Bike | Curve | Car stop | Guardrail | Color cone | Bump |
|---|---|---|---|---|---|---|---|---|---|---|
| ConvNeXt-T [17] | IM-RGB | 51.4 | 85.9 | 70.0 | 60.6 | 41.4 | 22.7 | 0.8 | 42.8 | 40.3 |
| | CSP-4 | **56.9** | **88.9** | **73.8** | **62.4** | **45.8** | **25.3** | **5.7** | **51.1** | **60.7** |
| Swin-T [18] | IM-RGB | 48.0 | 81.9 | 67.3 | 57.2 | 31.0 | 19.9 | **0.7** | 41.9 | 35.0 |
| | CSP-4 | **55.7** | **88.1** | **74.2** | **65.4** | **46.1** | **20.5** | 0.2 | **52.6** | **55.7** |

**Table 7.** Semantic segmentation accuracies on PST900 test set (%)

| Networks | Pretrain | mIoU | IoU per class | | | | |
| | | | Background | Fire extinguisher | Backpack | Hand drill | Survivor |
|---|---|---|---|---|---|---|---|
| ConvNeXt-T [17] | IM-RGB | 73.20 | 99.22 | 66.68 | 75.16 | 60.09 | 64.85 |
| | CSP-4 | **84.22** | **99.48** | **80.84** | **86.04** | **80.24** | **74.52** |
| Swin-T [18] | IM-RGB | 66.47 | 99.02 | 63.70 | 73.55 | 43.90 | 52.19 |
| | CSP-4 | **83.11** | **99.40** | **79.14** | **84.21** | **84.36** | **68.45** |

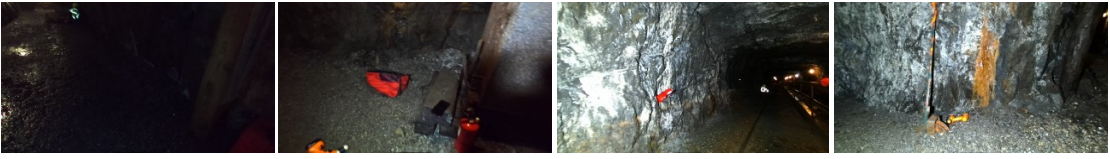

**Fig. 3.** Challenging scenes from PST900.

In the Potsdam dataset, we evaluated several spectral combinations: IR+R+G, R+G+B, R+G+B+IR, and R+G+B+IR+DSM. As summarized in Table 8, CSP pre-training consistently outperformed IM-RGB in achieving higher mF1 scores after fine-tuning across all channel configurations. Notably, CSP-3 slightly improved outcomes for R+G+B and IR+R+G, while CSP-4 and CSP-5 showed significant gains for R+G+B+IR and R+G+B+IR+DSM respectively, with CSP-4 boosting mF1 scores by up to 2.54% in ConvNeXt-T and 2.22% in Swin-T, and CSP-5 improving by 1.65% and 1.34% correspondingly. Overall, these results, detailed in Tables 5-8, underscore CSP's robust performance across datasets with varied channel types and combinations, enhancing semantic segmentation accuracy irrespective of the specific channel configuration in multi-channel datasets like Potsdam.

**Table 8.** Semantic segmentation accuracies on Potsdam test set (%)

| Networks | Pretrain | Feature Channel | mF1 | F1 score per class | | | | |
| --- | --- | --- | --- | --- | --- | --- | --- | --- |
| | | | | Impervious surface | Building | Low vegetation | Tree | Car |
| ConvNeXt-T [17] | IM-RGB | IR, R, G | 91.53 | 91.33 | 96.76 | **87.37** | 88.69 | 93.51 |
| | CSP-3 | IR, R, G | **91.73** | **91.58** | **97.00** | 87.29 | **89.00** | **93.79** |
| | IM-RGB | R, G, B | 91.47 | 91.86 | 96.66 | **86.60** | 88.80 | 93.42 |
| | CSP-3 | R, G, B | **91.66** | **92.04** | **96.75** | 86.39 | **89.09** | **94.02** |
| | IM-RGB | R, G, B, IR | 89.21 | 89.76 | 95.53 | 83.28 | 85.20 | 92.28 |
| | CSP-4 | R, G, B, IR | **91.75** | **91.57** | **96.82** | **87.37** | **88.73** | **94.27** |
| | IM-RGB | R, G, B, IR, | 90.23 | 91.12 | 96.83 | 85.67 | 86.68 | 90.84 |
| | CSP-5 | R, G, B, IR, | **91.88** | **91.44** | **97.32** | **87.75** | **89.00** | **93.90** |
| Swin-T [18] | IM-RGB | IR, R, G | 90.60 | **92.94** | 96.66 | 86.54 | 85.87 | 90.98 |
| | RSP [23] | IR, R, G | 90.03 | 92.65 | 96.35 | 86.02 | 85.39 | 89.75 |
| | CSP-3 | IR, R, G | **91.65** | 91.44 | **96.94** | **87.41** | **88.82** | **93.64** |
| | IM-RGB | R, G, B | 91.47 | 91.74 | **96.79** | 86.81 | 88.70 | 93.30 |
| | CSP-3 | R, G, B | **91.64** | **91.76** | 96.69 | **86.85** | **89.01** | **93.89** |
| | IM-RGB | R, G, B, IR | 89.47 | 90.06 | 95.63 | 84.21 | 85.63 | 91.79 |
| | CSP-4 | R, G, B, IR | **91.69** | **91.55** | **96.79** | **87.03** | **88.59** | **94.51** |
| | IM-RGB | R, G, B, IR, | 90.48 | 91.25 | 97.11 | 86.33 | 86.73 | 90.99 |
| | CSP-5 | R, G, B, IR, | **91.82** | **92.00** | **97.25** | **86.90** | **88.65** | **94.30** |

**B. Qualitative results.** To further assess the CSP's influence on semantic segmentation, we visually reviewed maps from networks trained with various strategies, with ConvNeXt-T model maps showcased in Fig. 4. In the iSAID scene (first column of Fig. 4), the baseline model only detected a small segment of the baseball diamond due to shadow-induced spectral perturbations. However, CSP-trained models identified a larger area of the diamond, suggesting CSP's effectiveness against shadow-related errors in segmentation, which are common issues [24]. These models also more successfully segmented a greater number of large vehicles. In the MFNet scene (second column of Fig. 4), both RGB and thermal baseline models missed bicycles, whereas CSP-trained models recognized most of these along with a person occluded by the Car Stop, overlooked even in manual labeling. For the PST900 scene (third column of Fig. 4), the baseline model struggled with segmenting the backpack and fire extinguisher due to similar colors and color variations. CSP-trained models showed improved segmentation accuracy. Lastly, in the Potsdam scene (final column of Fig. 4), the baseline inaccurately segmented a central impervious surface difficult to discern by the human eye. However, it was almost completely segmented by CSP-trained models, demonstrating CSP's potential in complex segmentation scenarios.

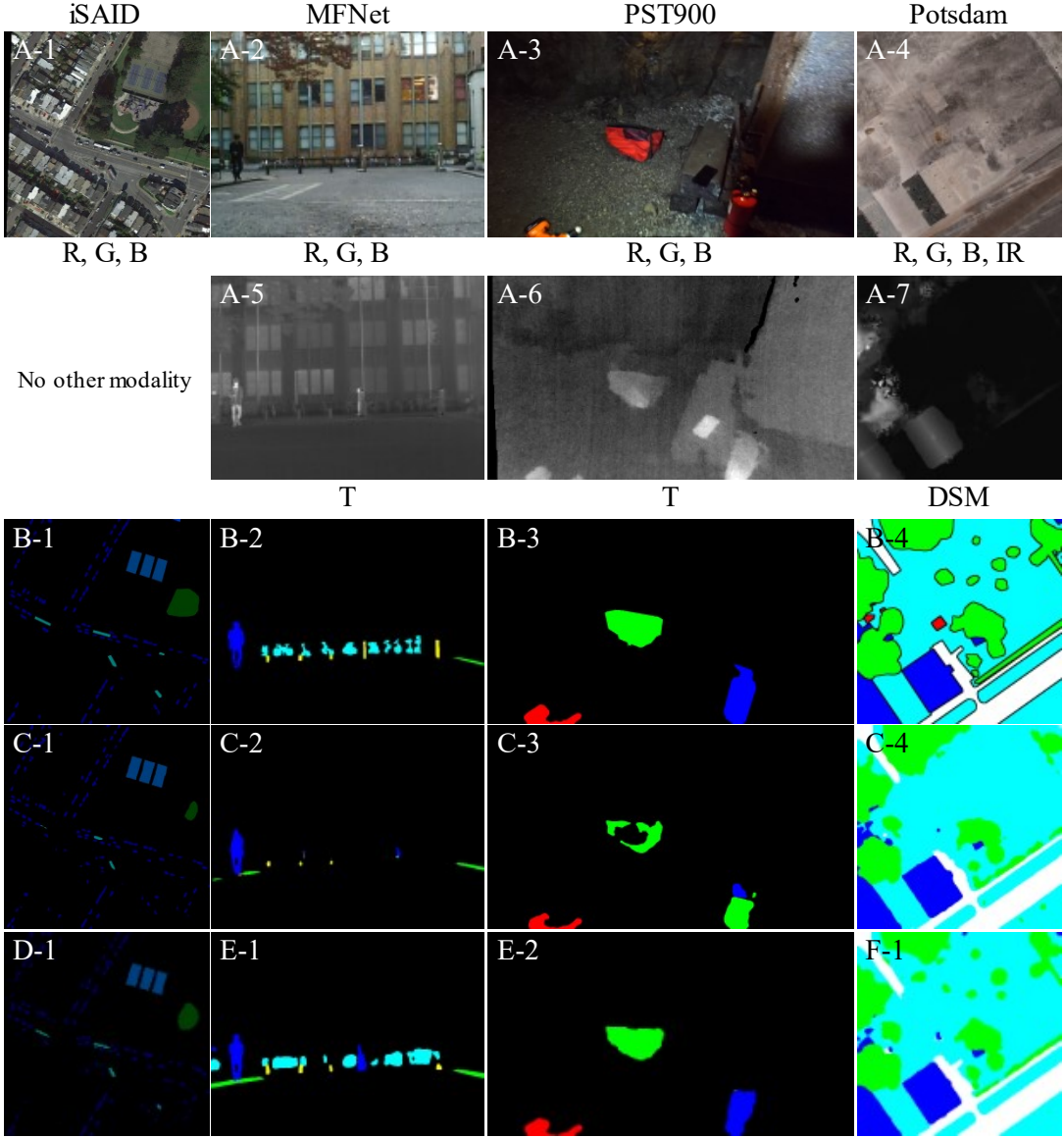

**Fig. 4.** Visual comparison of semantic segmentation maps from ConvNeXt-T models with distinct pre-training strategies: A-input images; B-ground-truth maps; C-segmentation by ImageNet pre-trained ConvNeXt-T; D-by ConvNeXt-T pre-trained with CSP-3; E-by CSP-4; F-by CSP-5.

## 5 Conclusion

In this study, we introduced the CSP strategy, a novel method to improve model generalization across diverse datasets by reducing dependence on spectral information during initial training. Experiments showed that CSP pre-trained on ImageNet alone achieved SOTA fine-tuning segmentation accuracies on several datasets: 67.4% mIoU on iSAID, 56.9% on MFNet, 84.22% on PST900, and a mean F1 score of 91.88% on Potsdam. These outcomes highlight CSP's ability to enhance model adaptability and effectiveness in various data modalities and environments. However, limitations were noted in precise edge detection, with blurred boundaries signaling a need for further research to combine CSP with edge-aware techniques for improved object boundary definition and accuracy.

Looking ahead, the simplicity and flexibility of the CSP methodology suggest promising avenues for expansion. Its compatibility with various network architectures and pre-training strategies makes it an excellent candidate for integration with contemporary techniques such as contrastive learning, masked image modeling, and multi-modal learning. Future work will explore these integrations to further validate and refine the effectiveness of CSP across broader computer vision tasks including object detection, human pose estimation, and super-resolution, that may benefit from robustness against spectral variations and enhanced spatial feature recognition. These directions could pave the way for developing more versatile and universally applicable visual recognition models, aligning with the needs of incrementally complex and diverse real-world applications.

**Acknowledgments.** This work was supported by the Xi'an Jiaotong-Liverpool University Research Enhancement Fund under Grant REF-21-01-003.

# References

1. Aryal, J., Sitaula, C., Aryal, S.: NDVI Threshold-Based Urban Green Space Mapping from Sentinel-2A at the Local Governmental Area (LGA) Level of Victoria, Australia. Land (Basel). 11, 351 (2022). https://doi.org/10.3390/land11030351.

2. Zhu, Q., Cai, Y., Fang, Y., Yang, Y., Chen, C., Fan, L., Nguyen, A.: Samba: Semantic segmentation of remotely sensed images with state space model. Heliyon. 10, e38495 (2024). https://doi.org/10.1016/j.heliyon.2024.e38495.

3. Zhu, Q., Fang, Y., Cai, Y., Chen, C., Fan, L.: Rethinking Scanning Strategies with Vision Mamba in Semantic Segmentation of Remote Sensing Imagery: An Experimental Study. IEEE J Sel Top Appl Earth Obs Remote Sens. 1–14 (2024). https://doi.org/10.1109/JSTARS.2024.3472296.

4. Rajbhandari, S., Aryal, J., Osborn, J., Lucieer, A., Musk, R.: Leveraging Machine Learning to Extend Ontology-Driven Geographic Object-Based Image Analysis (O-GEOBIA): A Case Study in Forest-Type Mapping. Remote Sens (Basel). 11, 503 (2019). https://doi.org/10.3390/rs11050503.

5. Cai, Y., Huang, H., Wang, K., Zhang, C., Fan, L., Guo, F.: Selecting Optimal Combination of Data Channels for Semantic Segmentation in City Information Modelling (CIM). Remote Sens (Basel). 13, 1367 (2021). https://doi.org/10.3390/rs13071367.

6. Cai, Y., Fan, L., Atkinson, P.M., Zhang, C.: Semantic Segmentation of Terrestrial Laser Scanning Point Clouds Using Locally Enhanced Image-Based Geometric Representations. IEEE Transactions on Geoscience and Remote Sensing. 60, 1–15 (2022). https://doi.org/10.1109/TGRS.2022.3161982.

7. Russakovsky, O., Deng, J., Su, H., Krause, J., Satheesh, S., Ma, S., Huang, Z., Karpathy, A., Khosla, A., Bernstein, M., Berg, A.C., Fei-Fei, L.: ImageNet Large Scale Visual Recognition Challenge. Int J Comput Vis. 115, 211–252 (2015). https://doi.org/10.1007/s11263-015-0816-y.

8. Hao, J., Chen, S.: Language-aware multiple datasets detection pretraining for DETRs. Neural Networks. 179, 106506 (2024). https://doi.org/10.1016/j.neunet.2024.106506.

9. Shirmard, H., Farahbakhsh, E., Müller, R.D., Chandra, R.: A review of machine learning in processing remote sensing data for mineral exploration. Remote Sens Environ. 268, 112750 (2022). https://doi.org/10.1016/j.rse.2021.112750.

10. Fang, Y., Cai, Y., Fan, L.: SDRCNN: A Single-Scale Dense Residual Connected Convolutional Neural Network for Pansharpening. IEEE J Sel Top Appl Earth Obs Remote Sens. 16, 6325–6338 (2023). https://doi.org/10.1109/JSTARS.2023.3292320.

11. Sun, X., Wang, P., Lu, W., Zhu, Z., Lu, X., He, Q., Li, J., Rong, X., Yang, Z., Chang, H., He, Q., Yang, G., Wang, R., Lu, J., Fu, K.: RingMo: A Remote Sensing Foundation Model With Masked Image Modeling. IEEE Transactions on Geoscience and Remote Sensing. 61, 1–22 (2023). https://doi.org/10.1109/TGRS.2022.3194732.

12. Jing, L., Tian, Y.: Self-Supervised Visual Feature Learning With Deep Neural Networks: A Survey. IEEE Trans Pattern Anal Mach Intell. 43, 4037–4058 (2021). https://doi.org/10.1109/TPAMI.2020.2992393.

13. Cai, Y., Aryal, J., Fang, Y., Huang, H., Fan, L.: OSTA: One-shot Task-adaptive Channel Selection for Semantic Segmentation of Multichannel Images. (2023).

14. Waqas Zamir, S., Arora, A., Gupta, A., Khan, S., Sun, G., Shahbaz Khan, F., Zhu, F., Shao, L., Xia, G.-S., Bai, X.: iSAID: A Large-scale Dataset for Instance Segmentation in Aerial Images, (2019).

15. Ha, Q., Watanabe, K., Karasawa, T., Ushiku, Y., Harada, T.: MFNet: Towards real-time semantic segmentation for autonomous vehicles with multi-spectral scenes. In: 2017 IEEE/RSJ International Conference on Intelligent Robots and Systems (IROS). pp. 5108–5115. IEEE (2017). https://doi.org/10.1109/IROS.2017.8206396.

16. Shivakumar, S.S., Rodrigues, N., Zhou, A., Miller, I.D., Kumar, V., Taylor, C.J.: PST900: RGB-Thermal Calibration, Dataset and Segmentation Network. In: 2020 IEEE International Conference on Robotics and Automation (ICRA). pp. 9441–9447. IEEE (2020). https://doi.org/10.1109/ICRA40945.2020.9196831.

17. Cai, Y., Fan, L., Fang, Y.: SBSS: Stacking-Based Semantic Segmentation Framework for Very High-Resolution Remote Sensing Image. IEEE Transactions on Geoscience and Remote Sensing. 61, 1–14 (2023). https://doi.org/10.1109/TGRS.2023.3234549.

18. Liu, Z., Lin, Y., Cao, Y., Hu, H., Wei, Y., Zhang, Z., Lin, S., Guo, B.: Swin Transformer: Hierarchical Vision Transformer using Shifted Windows. In: 2021 IEEE/CVF International Conference on Computer Vision (ICCV). pp. 9992–10002. IEEE (2021). https://doi.org/10.1109/ICCV48922.2021.00986.

19. Liu, Z., Mao, H., Wu, C.-Y., Feichtenhofer, C., Darrell, T., Xie, S.: A ConvNet for the 2020s. In: 2022 IEEE/CVF Conference on Computer Vision and Pattern Recognition (CVPR). pp. 11966–11976. IEEE (2022). https://doi.org/10.1109/CVPR52688.2022.01167.

20. Xiao, T., Liu, Y., Zhou, B., Jiang, Y., Sun, J.: Unified Perceptual Parsing for Scene Understanding, (2018).

21. Muhtar, D., Zhang, X., Xiao, P., Li, Z., Gu, F.: CMID: A Unified Self-Supervised Learning Framework for Remote Sensing Image Understanding. IEEE Transactions on Geoscience and Remote Sensing. 61, 1–17 (2023). https://doi.org/10.1109/TGRS.2023.3268232.

22. Zhu, Q., Fan, L., Weng, N.: Advancements in point cloud data augmentation for deep learning: A survey. Pattern Recognit. 153, 110532 (2024). https://doi.org/10.1016/j.patcog.2024.110532.

23. Wang, D., Zhang, J., Du, B., Xia, G.-S., Tao, D.: An Empirical Study of Remote Sensing Pretraining. IEEE Transactions on Geoscience and Remote Sensing. 61, 1–20 (2023). https://doi.org/10.1109/TGRS.2022.3176603.

24. Cai, Y., Fan, L., Zhang, C.: Semantic Segmentation of Multispectral Images via Linear Compression of Bands: An Experiment Using RIT-18. Remote Sens (Basel). 14, 2673 (2022). https://doi.org/10.3390/rs14112673.